\documentclass[journal]{IEEEtran}

\usepackage{cite}

\ifCLASSINFOpdf
   \usepackage[pdftex]{graphicx}
\else
\fi

\begin{document}

\title{Deep Representation Learning for Clustering of Health Tweets}

\author{Oguzhan~Gencoglu
\thanks{O. Gencoglu is with Faculty of Medicine and Health Technology, Tampere University, Tampere, 33014, Finland e-mail: (oguzhangencoglu90@gmail.com).}
}

\maketitle

\begin{abstract}
Twitter has been a prominent social media platform for mining population-level health data and accurate clustering of health-related tweets into topics is important for extracting relevant health insights. In this work, we propose deep convolutional autoencoders for learning compact representations of health-related tweets, further to be employed in clustering. We compare our method to several conventional tweet representation methods including bag-of-words, term frequency-inverse document frequency, Latent Dirichlet Allocation and Non-negative Matrix Factorization with 3 different clustering algorithms. Our results show that the clustering performance using proposed representation learning scheme significantly outperforms that of conventional methods for all experiments of different number of clusters. In addition, we propose a constraint on the learned representations during the neural network training in order to further enhance the clustering performance. All in all, this study introduces utilization of deep neural network-based architectures, i.e., deep convolutional autoencoders, for learning informative representations of health-related tweets.
\end{abstract}

\IEEEpeerreviewmaketitle

\section{Introduction}

Social media plays an important role in health informatics and Twitter has been one of the most influential social media channel for mining population-level health insights~\cite{vance2009social,paul2011you,Twitter}. These insights range from forecasting of influenza epidemics~\cite{achrekar2011predicting} to predicting adverse drug reactions~\cite{bian2012towards}. A notable challenge due to the short length of Twitter messages is categorization of tweets into topics in a supervised manner, i.e., topic classification, as well as in an unsupervised manner, i.e., clustering.

Classification of tweets into topics has been studied extensively~\cite{sriram2010short,batool2013precise,theodotou2015system}. Even though text classification algorithms can reach significant accuracy levels, supervised machine learning approaches require annotated data, i.e, topic categories to learn from for classification. On the other hand, annotated data is not always available as the annotation process is burdensome and time-consuming. In addition, discussions in social media evolve rapidly with recent trends, rendering Twitter a dynamic environment with ever-changing topics. Therefore, unsupervised approaches are essential for mining health-related information from Twitter.

Proposed methods for clustering tweets employ conventional text clustering pipelines involving preprocessing applied to raw text strings, followed by feature extraction which is then followed by a clustering algorithm~\cite{lu2013health,rosa2011topical,kaleel2015cluster}. Performance of such approaches depend highly on feature extraction in which careful engineering and domain knowledge is required~\cite{lecun2015deep}. Recent advancements in machine learning research, i.e., deep neural networks, enable efficient representation learning from raw data in a hierarchical manner~\cite{hinton2006reducing,bengio2013representation}. Several natural language processing (NLP) tasks involving Twitter data have benefited from deep neural network-based approaches including sentiment classification of tweets~\cite{severyn2015unitn}, predicting potential suicide attempts from Twitter~\cite{benton2017multi} and simulating epidemics from Twitter~\cite{zhao2015simnest}.

In this work, we propose deep convolutional autoencoders (CAEs) for obtaining efficient representations of health-related tweets in an unsupervised manner. We validate our approach on a publicly available dataset from Twitter by comparing the performance of our approach and conventional feature extraction methods on 3 different clustering algorithms. Furthermore, we propose a constraint on the learned representations during neural network training in order to further improve the clustering performance. We show that the proposed deep neural network-based representation learning method outperforms conventional methods in terms of clustering performance in experiments of varying number of clusters.

\section{Related Work}

Devising efficient representations of tweets, i.e., features, for performing clustering has been studied extensively. Most frequently used features for representing the text in tweets as numerical vectors are \textit{bag-of-words} (BoWs) and \textit{term frequency-inverse document frequency} (tf-idf) features~\cite{ramage2010characterizing,rosa2011topical,kaleel2015cluster,yang2015gis,lo2017unsupervised}. Both of these feature extraction methods are based on word occurrence counts and eventually, result in a sparse (most elements being zero) document-term matrix. Proposed algorithms for clustering tweets into topics include variants of hierarchical, density-based and centroid-based clustering methods; k-means algorithm being the most frequently used one~\cite{rosa2011topical,lo2017unsupervised,ma2017extracting}.

Numerous works on topic modeling of tweets are available as well. Topic models are generative models, relying on the idea that a given tweet is a mixture of topics, where a topic is a probability distribution over words~\cite{steyvers2007probabilistic}. Even though the objective in topic modeling is slightly different than that of pure clustering, representing each tweet as a topic vector is essentially a way of dimensionality reduction or feature extraction and can further be followed by a clustering algorithm. Proposed topic modeling methods include conventional approaches or variants of them such as Latent Dirichlet Allocation (LDA)~\cite{blei2003latent,ramage2010characterizing,rosa2011topical,prier2011identifying,hannachi2012community,yang2014identifying,paul2014discovering,kaveri2017framework,lo2017unsupervised,karami2018characterizingDiet,karami2018characterizingTrans} and Non-negative Matrix Factorization (NMF)~\cite{yan2012clustering,yang2015gis}. Note that topic models such as LDA are based on the notion that words belonging to a topic are more likely to appear in the same document and do not assume a distance metric between discovered topics.

In contrary to abovementioned feature extraction methods which are not specific to representation of tweets but rather generic in natural language processing, various works propose custom feature extraction methods for certain health-related information retrieval tasks from Twitter. For instance, Lim et al. engineered sentiment analysis features to discover latent infectious diseases from Twitter~\cite{lim2017unsupervised}. In order to track public health condition trends from Twitter, specific features are proposed by Parker at al. employing Wikipedia article index, i.e., treating the retrieval of medically-related Wikipedia articles as an indicator of a health-related condition~\cite{parker2013framework}. Custom user similarity features calculated from tweets were also proposed for building a framework for recommending health-related topics~\cite{kaveri2017framework}.

The idea of learning effective representations from raw data using neural networks has been employed in numerous machine learning domains such as computer vision and natural language processing~\cite{hinton2006reducing,bengio2013representation}. The concept relies on the hierarchical, layer-wise architecture of neural networks in which the raw input data is encoded into informative representations of lower dimensions (representations of higher dimensions are possible as well) in a highly non-linear fashion. Autoencoders, Denoising Autoencoders, Convolutional Autoencoders, Sparse Autoencoders, Stacked Autoencoders and combinations of these, e.g., Denoising Convolutional Autoencoders, are the most common deep neural network architectures specifically used for representation learning. In an autoencoder training, the network tries to reconstruct the input data at its output, which forces the model to capture the most salient features of the data at its intermediate layers. If the intermediate layers correspond to a lower dimensional latent space than the original input, such autoencoders are also known as \textit{undercomplete}. Activations extracted from these layers can be considered as compact, non-linear representations of the input.

Another significant advancement in neural network-based representation learning in NLP tasks is \textit{word embeddings} (also called \textit{distributed representation of words}). By representing each word in a given vocabulary with a real-valued vector of a fixed dimension, word embeddings enable capturing of lexical, semantic or even syntactic similarities between words. Typically, these vector representations are learned from large corpora and can be used to enhance the performance of numerous NLP tasks such as document classification, question answering and machine translation. Most frequently used word embeddings are \textit{word2vec}~\cite{mikolov2013efficient} and \textit{GloVe} (Global Vectors for Word Representation)~\cite{pennington2014glove}. Both of these are extracted in an unsupervised manner and are based on the \textit{distributional hypothesis}~\cite{harris1954distributional}, i.e., the assumption that words that occur in the same contexts tend to have similar meanings. Both word2vec and GloVe treat a word as a smallest entity to train on. A shift in this paradigm was introduced by \textit{fastText}~\cite{bojanowski2016enriching}, which treats each word as a bag of character n-grams. Consequently, fastText embeddings are shown to have better representations for rare words~\cite{bojanowski2016enriching}. In addition, one can still construct a vector representation for an out-of-vocabulary word which is not possible with word2vec or GloVe embeddings~\cite{bojanowski2016enriching}. Enhanced methods for deducting better word and/or sentence representations were recently introduced as well by Peters et al. with the name \textit{ELMo} (Embeddings from Language Models)~\cite{peters2018deep} and by Devlin et al. with the name \textit{BERT} (Bidirectional Encoder Representations from Transformers)~\cite{devlin2018bert}. All of these word embedding models are trained on large corpora such as Wikipedia, in an unsupervised manner. For analyzing tweets, word2vec and GloVe word embeddings have been employed for topical clustering of tweets~\cite{dai2017social}, topic modeling~\cite{sridhar2015unsupervised,nguyen2018improving} and extracting depression symptoms from tweets~\cite{ma2017extracting}.

Metrics for evaluating the performance of clustering algorithms varies depending on whether the ground truth topic categories are available or not. If so, frequently used metrics are \textit{accuracy} and \textit{normalized mutual information}. In the case of absence of ground truth labels, one has to use internal clustering criterions such as Calinski-Harabasz (CH) score~\cite{calinski1974dendrite} and Davies-Bouldin index~\cite{davies1979cluster}. Arbelaitz et al. provides an extensive comparative study of cluster validity indices~\cite{arbelaitz2013extensive}. 

\section{Methods}

\subsection{Dataset}

For this study, a publicly available dataset is used~\cite{uciML}. The dataset consisting of tweets has been collected using Twitter API and was initially introduced by Karami et al.~\cite{karami2018fuzzy}. Earliest tweet dates back to 13 June 2011 where the latest one has a timestamp of 9 April 2015. The dataset consists of 63,326 tweets in English language, collected from Twitter channels of 16 major health news agencies. List of health news channels and the number of tweets in the dataset from each channel can be examined from Table~\ref{table1}.

\begin{table}[htbp]
\centering
  \caption{Number of tweets, total number of words, number of unique words and average number of words for tweets from 16 health-related twitter channels.}
  \begin{tabular}{p{2.40cm}p{1.1cm}p{1.15cm}p{1.3cm}p{0.8cm}}
     \hline
     \normalfont Twitter Channel & Number of Tweets & Number of Words & Number of Unique Words &  Mean Word Count \\
     \hline
     BBC Health & 3,929 & 22,543 & 4,334 & 5.7  \\
     CBC Health & 3,741 & 34,144 & 6,529 & 9.1\\
     CNN Health & 4,061 & 45,369 & 6,568 & 11.2 \\
     Everyday Health & 3,239 & 37,032 & 3,966 & 11.4  \\
     Fox News Health & 2,000 & 18,134 & 4,315 & 9.1  \\
     Guardian Healthcare & 2,997 & 42,481 & 4,139 & 14.2  \\
     Goodhealth & 7,864 & 105,370 & 8,002 & 13.4  \\
     Kaiser Health & 3,509 & 39,182 & 5,133 & 11.2  \\
     LA Times Health & 4,171 & 50,715 & 7,648 & 12.2  \\
     MSN Health & 3,199 & 26,252 & 4,275 & 8.2  \\
     NBC Health & 4,215 & 35,909 & 5,910 & 8.5  \\
     NPR Health & 4,837 & 43,427 & 7,303 & 9.0 \\
     NY Times Health & 6,245 & 62,726 & 8,567 & 10.0 \\
     Reuters Health & 4,719 & 44,210 & 6,482 & 9.4  \\
     US News Health & 1,400 & 16,546 & 2,869 & 11.8 \\
     WSJ Health & 3,200 & 40,317 & 6,792 & 12.6  \\
     \hline
  \end{tabular}
  \label{table1}
\end{table}

\begin{figure}
      \centering
          \includegraphics[width=0.948\columnwidth,trim={8 8.0cm 8 0},clip]{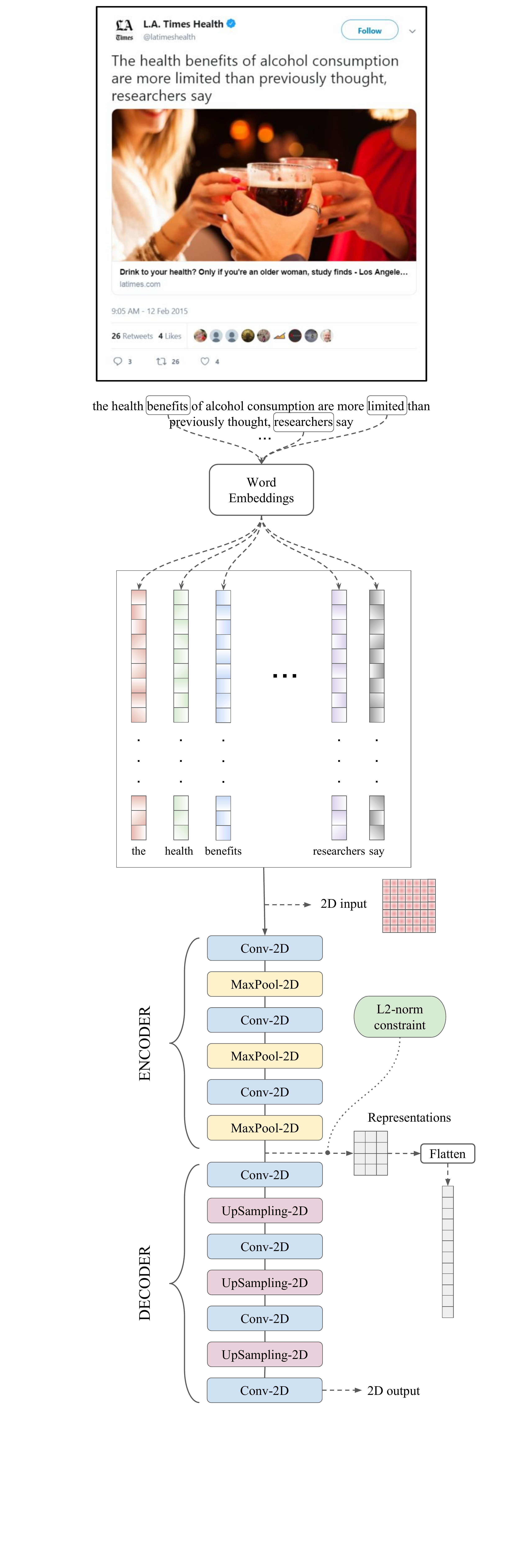}
      \caption{Proposed representation learning method depicting the overall flow starting from a tweet to the learned features, including the architecture of the convolutional autoencoder.}
      \label{figure1}
\end{figure}

The outlook of a typical tweet from the dataset can be examined from Figure~\ref{figure1}. For every tweet, the raw data consists of the tweet text and in most cases followed by a url to the original news article of the particular news source. This url string, if available, is removed from each tweet as it does not possess any natural language information. As Twitter allows several ways for users to interact such as \textit{retweeting} or \textit{mentioning}, these actions appear in the raw text as well. For retweets, an indicator string of "RT" appears as a prefix in the raw data and for user mentions, a string of form "@username" appears in the raw data. These two tokens are removed as well. In addition, hashtags are converted to plain tokens by removal of the "\#" sign appearing before them (e.g. \textless \#pregnancy\textgreater~becomes \textless pregnancy\textgreater). Number of words, number of unique words and mean word counts for each Twitter channel can also be examined from Table~\ref{table1}. Longest tweet consists of 27 words.

\subsection{Conventional Representations}

For representing tweets, 5 conventional representation methods are proposed as baselines. 

\begin{enumerate}
\item \textit{Word frequency features}: For word occurrence-based representations of tweets, conventional tf-idf and BoWs are used to obtain the \textit{document-term matrix} of $N \times P$ in which each row corresponds to a tweet and each column corresponds to a unique word/token, i.e., $N$ data points and $P$ features. As the document-term matrix obtained from tf-idf or BoWs features is extremely sparse and consequently redundant across many dimensions, dimensionality reduction and topic modeling to a lower dimensional latent space is performed by the methods below.
\item \textit{Principal Component Analysis (PCA)}: PCA is used to map the word frequency representations from the original feature space to a lower dimensional feature space by an \textit{orthogonal linear transformation} in such a way that the first principal component has the highest possible variance and similarly, each succeeding component has the highest variance possible while being orthogonal to the preceding components. Our PCA implementation has a time complexity of $\mathcal{O}(NP^2 + P^3)$.
\item \textit{Truncated Singular Value Decomposition (t-SVD)}: Standard SVD and t-SVD are commonly employed dimensionality reduction techniques in which a matrix is reduced or approximated into a \textit{low-rank} decomposition. Time complexity of SVD and t-SVD for $S$ components are $\mathcal{O}(min(NP^2, N^2P))$ and $\mathcal{O}(N^2S)$, respectively (depending on the implementation). Contrary to PCA, t-SVD can be applied to sparse matrices efficiently as it does not require data normalization. When the data matrix is obtained by BoWs or tf-idf representations as in our case, the technique is also known as \textit{Latent Semantic Analysis}.
\item \textit{LDA}: Our LDA implementation employs online variational Bayes algorithm introduced by Hoffman et al. which uses stochastic optimization to maximize the objective function for the topic model~\cite{hoffman2010online}.
\item \textit{NMF}: As NMF finds two non-negative matrices whose product approximates the non-negative document-term matrix, it allows regularization. Our implementation did not employ any regularization and the divergence function is set to be squared error, i.e., Frobenius norm.
\end{enumerate}

\subsection{Representation Learning}

We propose 2D convolutional autoencoders for extracting compact representations of tweets from their raw form in a highly non-linear fashion. In order to turn a given tweet into a 2D structure to be fed into the CAE, we extract the word vectors of each word using word embedding models, i.e., for a given tweet, $t$, consisting of $W$ words, the 2D input is $I_{t} \in \R^{W \times D}$ where $D$ is the embedding vector dimension. We compare 4 different word embeddings namely \textit{word2vec}, \textit{GloVe}, \textit{fastText} and \textit{BERT} with embedding vector dimensions of 300, 300, 300 and 768, respectively. We set the maximum sequence length to 32, i.e., for tweets having less number of words, the input matrix is padded with zeros. As word2vec and GloVe embeddings can not handle out-of-vocabulary words, such cases are represented as a vector of zeros. The process of extracting word vector representations of a tweet to form the 2D input matrix can be examined from Figure~\ref{figure1}. 

The CAE architecture can be considered as consisting of 2 parts, ie., the \textit{encoder} and the \textit{decoder}. The encoder, $f_{enc}(\cdot)$, is the part of the network that compresses the input, $I$, into a latent space representation, $U$, and the decoder, $f_{dec}(\cdot)$ aims to reconstruct the input from the latent space representation (see equation~\ref{eq1}). In essence,

\begin{equation}
U = f_{enc}(I) = f_{L}(f_{L-1}(...f_{1}(I)))
\label{eq1}
\end{equation}
where $L$ is the number of layers in the encoder part of the CAE.

The encoder in the proposed architecture consists of three 2D convolutional layers with 64, 32 and 1 filters, respectively. The decoder follows the same symmetry with three convolutional layers with 1, 32 and 64 filters, respectively and an output convolutional layer of a single filter (see Figure~\ref{figure1}). All convolutional layers have a kernel size of (3$\times$3) and an activation function of \textit{Rectified Linear Unit} (ReLU) except the output layer which employs a \textit{linear} activation function. Each convolutional layer in the encoder is followed by a \textit{2D MaxPooling} layer and similarly each convolutional layer in the decoder is followed by a \textit{2D UpSampling} layer, serving as an inverse operation (having the same parameters). The pooling sizes for pooling layers are (2$\times$5), (2$\times$5) and (2$\times$2), respectively for the architectures when word2vec, GloVe and fastText embeddings are employed. With this configuration, an input tweet of size $32 \times 300$ (corresponding to maximum sequence length $\times$ embedding dimension, $D$) is downsampled to size of $4 \times 6$ out of the encoder (bottleneck layer). As BERT word embeddings have word vectors of fixed size 768, the pooling layer sizes are chosen to be (2$\times$8), (2$\times$8) and (2$\times$2), respectively for that case. In summary, a representation of $4 \times 6 = 24$ values is learned for each tweet through the encoder, e.g., for fastText embeddings the flow of dimensions after each encoder block is as such : $32 \times 300 \rightarrow 16 \times 60 \rightarrow 8 \times 12 \rightarrow 4 \times 6$.

In numerous NLP tasks, an \textit{Embedding Layer} is employed as the first layer of the neural network which can be initialized with the word embedding matrix in order to incorporate the embedding process into the architecture itself instead of manual extraction. In our case, this was not possible because of nonexistence of an inversed embedding layer in the decoder (as in the relationship between \textit{MaxPooling} layers and \textit{UpSampling} layers) as an embedding layer is not differentiable.

Training of autoencoders tries to minimize the reconstruction error/loss, i.e., the deviation of the reconstructed output from the input. $L_2$-loss or \textit{mean square error} (MSE) is chosen to be the loss function. In autoencoders, minimizing the $L_2$-loss is equivalent to maximizing the \textit{mutual information} between the reconstructed inputs and the original ones~\cite{vincent2010stacked}. In addition, from a probabilistic point of view, minimizing the $L_2$-loss is the same as maximizing the probability of the parameters given the data, corresponding to a \textit{maximum likelihood estimator}. The optimizer for the autoencoder training is chosen to be Adam due to its faster convergence abilities~\cite{kingma2014adam}. The learning rate for the optimizer is set to $10^{-5}$ and the batch size for the training is set to 32. Random split of 80\% training-20\% validation set is performed for monitoring convergence. Maximum number of training epochs is set to 50.

\subsection{$L_2$-norm Constrained Representation Learning}

Certain constraints on neural network weights are commonly employed during training in order to reduce overfitting, also known as \textit{regularization}. Such constraints include $L_1$ regularization, $L_2$ regularization, orthogonal regularization etc. Even though regularization is a common practice, standard training of neural networks do not inherently impose any constraints on the learned representations (activations), $U$, other than the ones compelled by the activation functions (e.g. ReLUs resulting in non-negative outputs). Recent advancements in computer vision research show that constraining the learned representations can enhance the effectiveness of representation learning, consequently increasing the clustering performance~\cite{huang2014deep,aytekin2018clustering}.

\begin{equation}
\begin{aligned}
& \text{minimize}
& & L = 1/_N \left\lVert I - f_{dec}(f_{enc}(I))\right\rVert^2_{2} \\
& \text{subject to}
& & \left\lVert f_{enc}(I)\right\rVert^2_{2} = 1
\end{aligned}
\label{eq2}
\end{equation}

We propose an $L_2$ norm constraint on the learned representations out of the bottleneck layer, $U$. Essentially, this is a hard constraint introduced during neural network training that results in learned features with unit $L_2$ norm out of the bottleneck layer (see equation~\ref{eq2} where $N$ is the number of data points). Training a deep convolutional autoencoder with such a constraint is shown to be much more effective for image data than applying $L_2$ normalization on the learned representations after training~\cite{aytekin2018clustering}. To the best of our knowledge, this is the first study to incorporate $L_2$ norm constraint in a task involving text data.

\subsection{Evaluation}


In order to fairly compare and evaluate the proposed methods in terms of effectiveness in representation of tweets, we fix the number of features to 24 for all methods and feed these representations as an input to 3 different clustering algorithms namely, \textit{k-means}, \textit{Ward} and \textit{spectral clustering} with cluster numbers of 10, 20 and 50. Distance metric for k-means clustering is chosen to be \textit{euclidean} and the linkage criteria for Ward clustering is chosen to be minimizing the sum of differences within all clusters, i.e., recursively merging pairs of clusters that minimally increases the within-cluster variance in a hierarchical manner. For spectral clustering, Gaussian kernel has been employed for constructing the affinity matrix. We also run experiments with tf-idf and BoWs representations without further dimensionality reduction as well as concatenation of all word embeddings into a long feature vector. For evaluation of clustering performance, we use Calinski-Harabasz score~\cite{calinski1974dendrite}, also known as the \textit{variance ratio criterion}. CH score is defined as the ratio between the within-cluster dispersion and the between-cluster dispersion. CH score has a range of $[0, +\infty]$ and a higher CH score corresponds to a better clustering. Computational complexity of calculating CH score is $\mathcal{O}(N)$.

For a given dataset $X$ consisting of $N$ data points, i.e., $X = \big\{x_1, x_2, ... , x_N\big\}$ and a given set of disjoint clusters $C$ with $K$ clusters, i.e., $C = \big\{c_1, c_2, ... , c_K\big\}$, Calinski-Harabasz score, $S_{CH}$, is defined as
\begin{equation}
S_{CH} = \frac{N-K}{K-1}\frac{\sum_{c_k \in C}^{}{N_k \left\lVert \overline{c_k}-\overline{X}\right\rVert^2_{2}}}{\sum_{c_k \in C}^{}{}\sum_{x_i \in c_k}^{}{\left\lVert x_i-\overline{c_k}\right\rVert^2_{2}}}
\end{equation}
where $N_k$ is the number of points belonging to the cluster $c_k$, $\overline{X}$ is the centroid of the entire dataset, $\frac{1}{N}\sum_{x_i \in X}{x_i}$ and $\overline{c_k}$ is the centroid of the cluster $c_k$, $\frac{1}{N_k}\sum_{x_i \in c_k}{x_i}$. 

For visual validation, we plot and inspect the t-Distributed Stochastic Neighbor Embedding (t-SNE)~\cite{maaten2008visualizing} and Uniform Manifold Approximation and Projection (UMAP)~\cite{mcinnes2018umap} mappings of the learned representations as well. Implementation of this study is done in Python (version 3.6) using \textit{scikit-learn} and \textit{TensorFlow} libraries~\cite{pedregosa2011scikit,abadi2016tensorflow} on a 64-bit Ubuntu 16.04 workstation with 128 GB RAM. Training of autoencoders are performed with a single NVIDIA Titan Xp GPU.

\begin{figure}
      \centering
          \includegraphics[width=1.0\columnwidth,trim={0 0cm 0 0},clip]{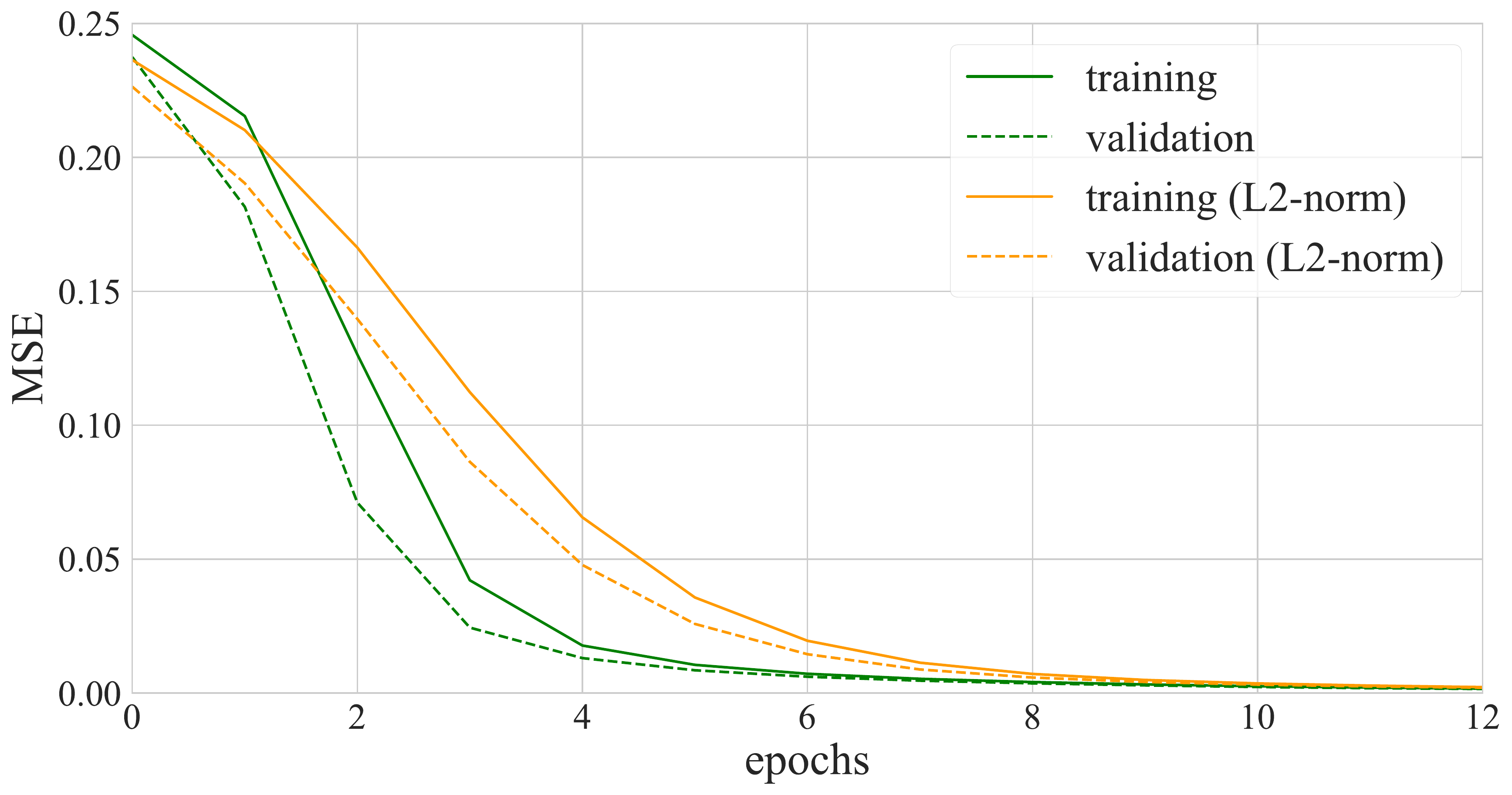}
      \caption{Learning curves depicting training and validation losses of CAE and $L_2$-norm constrained CAE architectures for fastText embeddings.}
      \label{figure2}
\end{figure}
\begin{figure*}[!htbp] 
      \centering
          \includegraphics[width=2.0\columnwidth,trim={0 0cm 0 0},clip]{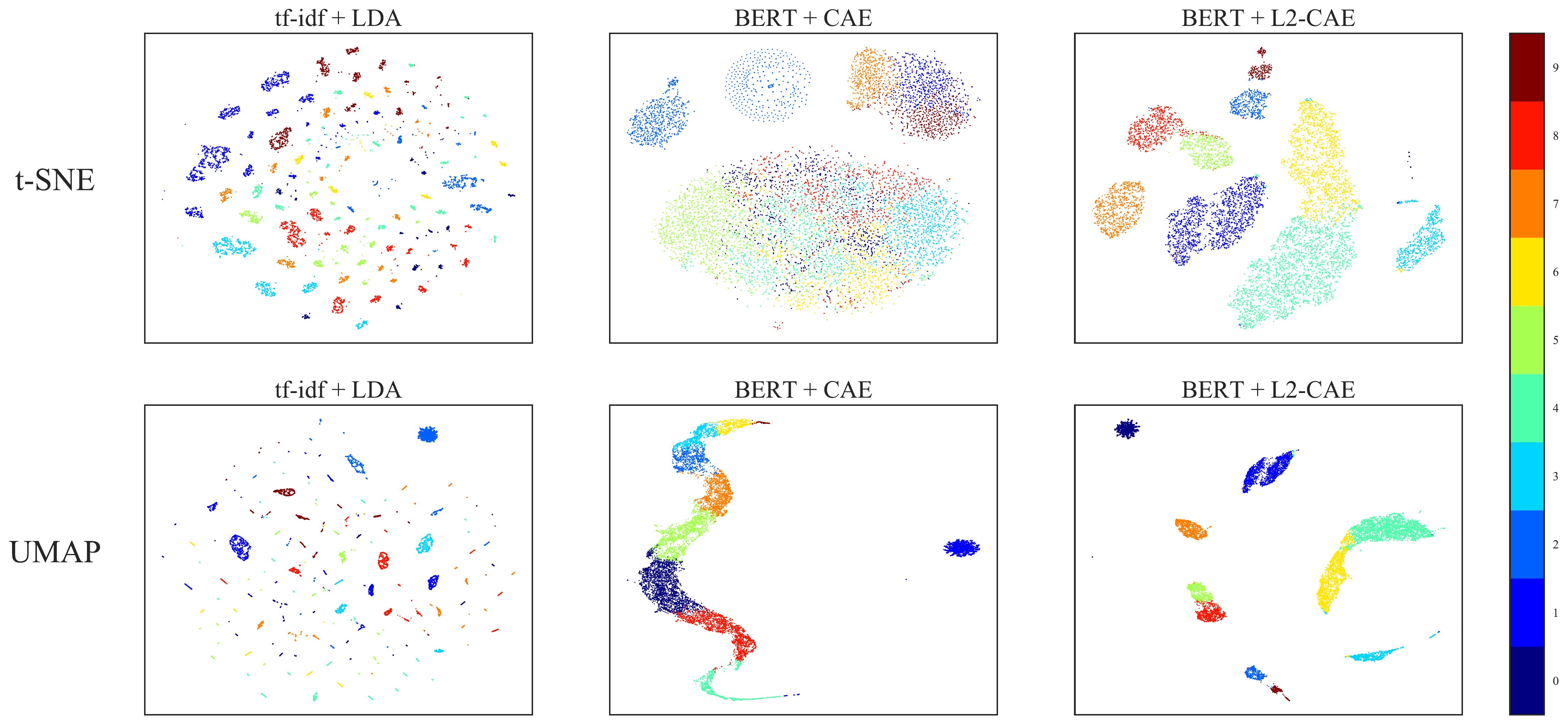}
      \caption{UMAP and t-SNE visualizations of representations extracted by LDA, CAE and $L_2$-norm constrained CAE (each having a length of 24) and coloring based on k-means clustering of the representations into 10 clusters.}
      \label{figure3}
\end{figure*}

\section{Results}

Performance of the representations tested on 3 different clustering algorithms, i.e., CH scores, for 3 different cluster numbers can be examined from Table~\ref{table2}. $L_2$-norm constrained CAE is simply referred as $L_2$-CAE in Table~\ref{table2}. Same table shows the number of features used for each method as well. Document-term matrix extracted by BoWs and tf-idf features result in a sparse matrix of $63,326 \times 13,026$ with a sparsity of 0.9994733. Similarly, concatenation of word embeddings result in a high number of features with $32 \times 300 = 9,600$ for word2vec, GloVe and fastText, $32 \times 768 = 24,576$ for BERT embeddings. In summary, the proposed method of learning representations of tweets with CAEs outperform all of the conventional algorithms. When representations are compared with Hotelling's $T^2$ test (multivariate version of $t$-test), every representation distribution learned by CAEs are shown to be statistically significantly different than every other conventional representation distribution with $p<0.001$. In addition, introducing the $L_2$-norm constraint on the learned representations during training enhances the clustering performance further (again $p<0.001$ when comparing for example fastText+CAE vs. fastText+$L_2$-CAE). An example learning curve for CAE and $L_2$-CAE with fastText embeddings as input can also be seen in Figure~\ref{figure2}.

\begin{table*}[!htbp] 
\centering
  \caption{Calinski-Harabasz scores for several conventional methods and proposed cae-based methods for 3 different clustering algorithms and 3 different number of clusters.}
  \begin{tabular}{|p{2.9cm}|p{1.2cm}|p{1.0cm}p{0.7cm}p{1.1cm}|p{1.0cm}p{0.7cm}p{1.1cm}|p{1.0cm}p{0.7cm}p{1.1cm}|}
     \cline{3-11}
     \multicolumn{2}{l}{} & \multicolumn{3}{|c|}{ 10 clusters} & \multicolumn{3}{|c|}{ 20 clusters}& \multicolumn{3}{|c|}{ 50 clusters}\\
     \hline
     \normalfont \hfil Tweet Representation & \hfil Features & k-means & Ward & spectral & k-means & Ward & spectral & k-means & Ward & spectral\\
     \hline
     tf-idf & \hfil 13026 & 23 & 14 & 9 & 15 & 12 & 7 & 12 & 9 & 5 \\
     BoW & \hfil 13026 & 23 & 25 & 9 & 18 & 19 & 7 & 15 & 13 & 5 \\
     word2vec (concatenated) & \hfil 9600 & 174 & 152 & 110 & 78 & 61 & 64 & 41 & 33 & 21 \\
     GloVe (concatenated) & \hfil 9600 & 157 & 131 & 119 & 97 & 63 & 71 & 43 & 37 & 27 \\
     fastText (concatenated) & \hfil 9600 & 212 & 163 & 135 & 115 & 84 & 97 & 54 & 50 & 39 \\
     BERT (concatenated) &\hfil 24576 & 206 & 171 & 138 & 123 & 85 & 94 & 57 & 42 & 41 \\
     \hline
     tf-idf + PCA & \hfil 24 & 443 & 333 & 320 & 421 & 316 & 309 & 421 & 396 & 395 \\
     BoW + PCA & \hfil 24 & 679 & 411 & 679 & 543 & 412 & 498 & 507 & 489 & 495\\
     tf-idf + t-SVD & \hfil 24 & 419 & 294 & 284 & 437 & 326 & 337 & 404 & 384  & 394\\
     BoW + t-SVD & \hfil 24 & 633 & 385 & 630 & 579 & 427 & 510 & 495 & 480 & 493 \\
     tf-idf + LDA & \hfil 24 & 684 & 497 & 627 & 621 & 486 & 620 & 638 & 433 & 614\\
     BoW + LDA & \hfil 24 & 408 & 280 & 403 & 355 & 260 & 359 & 284 & 214 & 271 \\
     tf-idf + NMF & \hfil 24 & 444 & 406 & 434 & 413 & 452 & 449 & 521 & 527 & 594 \\
     BoW + NMF & \hfil 24 & 512 & 477 & 636 & 491 & 460 & 563 & 591 & 556 & 600 \\
     \hline
     word2vec + CAE & \hfil 24  & 1851 & 1570 & 1726  & 1492 & 1357 & 1387 & 1317 & 1241 & 1040\\
     GloVe + CAE & \hfil 24 & 1953 & 1612 & 1696 & 1499 & 1302 & 1278 & 1367 & 1278 & 1102\\
     fastText + CAE & \hfil 24 & 3520 & 3173 & 3297 & 1914 & 1699 & 1772 & 1765 & 1567 & 1677\\
     BERT + CAE & \hfil 24 & 3467 & 3203 & 3288 & 2032 & 1768 & 1882 & 1834 & 1645 & 1711 \\
     word2vec + $L_2$-CAE & \hfil 24 & 3060 & 2964 & 2998 & 2513 & 2424 & 2501 & 2284 & 2043 & 2193\\
     GloVe + $L_2$-CAE & \hfil 24 & 3100 & 2931 & 3017 & 2602 & 2499 & 2526 & 2280 & 2076 & 2200 \\
     fastText + $L_2$-CAE & \hfil 24 & 6894 & 6884 & 6803 & \textbf{5839} & 5684 & 5743 & \textbf{4304} & 4187 & 4016 \\
     BERT + $L_2$-CAE & \hfil 24 & \textbf{7703} & 7071 & 6972 & 5768 & 5606 & 4554 & 4172 & 4014 & 2559 \\
     \hline
  \end{tabular}
  \label{table2}
\end{table*}

Detailed inspection of tweets that are clustered into the same cluster as well as visual analysis of the formed clusters is also performed. Figure~\ref{figure3} shows the t-SNE and UMAP mappings (onto 2D plane) of the 10 clusters formed by k-means algorithm for LDA, CAE and $L_2$-CAE representations. Below are several examples of tweets sampled from one of the clusters formed by k-means in the 50 clusters case (fastText embeddings fed into $L_2$-CAE):
\begin{itemize}
\item{\textless Suicide risk falls after talk therapy\textgreater}
\item{\textless Air pollution may be tied to anxiety\textgreater}
\item{\textless Stress, depression boost risks for heart patients\textgreater}
\item{\textless Nearly 1 in 5 Americans who has been out of work for at least 1 year is clinically depressed.\textgreater}
\item{\textless Study shows how exercise protects the brain against depression\textgreater}
\end{itemize}

\section{Discussion}

Overall, we show that deep convolutional autoencoder-based feature extraction, i.e., representation learning, from health related tweets significantly enhances the performance of clustering algorithms when compared to conventional text feature extraction and topic modeling methods (see Table~\ref{table2}). This statement holds true for 3 different clustering algorithms (k-means, Ward, spectral) as well as for 3 different number of clusters. In addition, proposed constrained training ($L_2$-norm constraint) is shown to further improve the clustering performance in each experiment as well (see Table~\ref{table2}). A Calinski-Harabasz score of 4,304 has been achieved with constrained representation learning by CAE for the experiment of 50 clusters formed by k-means clustering. The highest CH score achieved in the same experiment setting by conventional algorithms was 638 which was achieved by LDA applied of tf-idf features.

Visualizations of t-SNE and UMAP mappings in Figure~\ref{figure3} show that $L_2$-norm constrained training results in higher separability of clusters. The benefit of this constraint is especially significant in the performance of k-means clustering (see Table~\ref{table2}). This phenomena is not unexpected as k-means clustering is based on $L_2$ distance as well. The difference in learning curves for regular and constrained CAE trainings is also expected. Constrained CAE training converges to local minimum slightly later than unconstrained CAE, i.e., training of $L_2$-CAE is slightly slower than that of CAE due to the introduced contraint (see Figure~\ref{figure2}).

When it comes to comparison between word embeddings, fastText and BERT word vectors result in the highest CH scores whereas word2vec and GloVe embeddings result in significantly lower performance. This observation can be explained by the nature of word2vec and GloVe embeddings which can not handle out-of-vocabulary tokens. Numerous tweets include names of certain drugs which are more likely to be absent in the vocabulary of these models, consequently resulting in vectors of zeros as embeddings. However, fastText embeddings are based on character n-grams which enables handling of out-of-vocabulary tokens, e.g., fastText word vectors of the tokens \textless \textit{acetaminophen}\textgreater~and \textless \textit{paracetamol}\textgreater~are closer to each other simply due to shared character sequence, \textless \textit{acetam}\textgreater, even if one of them is not in the vocabulary. Note that, \textless \textit{acetaminophen}\textgreater~and \textless \textit{paracetamol}\textgreater~are different names for the same drug.

Using tf-idf or BoWs features directly results in very poor performance. Similarly, concatenating word embeddings to create thousands of features results in significantly low performance compared to methods that reduce these features to 24. The main reason is that the bias-variance trade-off is dominated by the bias in high dimensional settings especially in Euclidean spaces~\cite{friedman1997bias}. Due to very high number of features (relative to the number of observations), the radius of a given region varies with respect to the $n$th root of its volume, whereas the number of data points in the region varies roughly linearly with the volume~\cite{friedman1997bias}. This phenomena is known as \textit{curse of dimensionality}. As topic models such as LDA and NMF are designed to be used on documents that are sufficiently long to extract robust statistics from, extracted topic vectors fall short in performance as well when it comes to tweets due to short texts.

The main limitation of this study is the absence of topic labels in the dataset. As a result, internal clustering measure of Calinski-Harabasz score was used for evaluating the performance of the formed clusters instead of accuracy or normalized mutual information. Even though CH score is shown to be able to capture clusters of different densities and presence of subclusters, it has difficulties capturing highly noisy data and skewed distributions~\cite{liu2010understanding}. In addition, used clustering algorithms, i.e., k-means, Ward and spectral clustering, are hard clustering algorithms which results in non-overlapping clusters. However, a given tweet can have several topical labels.



Future work includes representation learning of health-related tweets using deep neural network architectures that can inherently learn the sequential nature of the textual data such as recurrent neural networks, e.g., Long Short-Term Memory (LSTM), Gated Recurrent Unit (GRU) etc. Sequence-to-sequence autoencoders are main examples of such architectures and they have been shown to be effective in encoding paragraphs from Wikipedia and other corpora to lower dimensions~\cite{li2015hierarchical}. Furthermore, encodings out of a bidirectional GRU will be tested for clustering performance, as such architectures have been employed to represent a given tweet in other studies~\cite{dhingra2016tweet2vec,wang2017hierarchical,vakulenko2017character}. 


\section{Conclusion}

In summary, we show that deep convolutional autoencoders can effectively learn compact representations of health-related tweets in an unsupervised manner. Conducted analysis show that the proposed representation learning scheme outperforms conventional feature extraction methods in three different clustering algorithms. In addition, we propose a constraint on the learned representation in order to further increase the clustering performance. Future work includes comparison of our model with recurrent neural architectures for clustering of health-related tweets. We believe this study serves as an advancement in the field of natural language processing for health informatics especially in clustering of short-text social media data.

\bibliographystyle{IEEEtran}
\bibliography{references}

\end{document}